\documentclass[letterpaper]{article}%
\usepackage[preprint]{aaai2027}%
\usepackage[hyphens]{url}%
\usepackage{graphicx}%
\usepackage{natbib}%
\usepackage{caption}%
\usepackage{booktabs}
\usepackage{array}
\usepackage{amsmath,amssymb,amsthm}
\usepackage{enumitem}
\usepackage{tikz}
\usetikzlibrary{arrows.meta,positioning,fit,shapes.geometric,decorations.pathreplacing,calc}
\usepackage[capitalise]{cleveref}
\crefname{figure}{Figure}{Figures}
\crefformat{equation}{(#2#1#3)}
\Crefformat{equation}{(#2#1#3)}

\definecolor{ptrue}{HTML}{CDE8C9}
\definecolor{pfalse}{HTML}{F4C7C3}
\definecolor{pcfill}{HTML}{CFE0F3}
\definecolor{pcline}{HTML}{2E5E8C}
\definecolor{relline}{HTML}{C8611B}
\definecolor{lockfill}{HTML}{E4ECF6}
\definecolor{deadgray}{HTML}{9A9A9A}
\tikzset{
  comp/.style={draw, rounded corners=3pt, fill=black!7, minimum width=7.5mm, minimum height=5.5mm, inner sep=1.5pt, font=\small},
  prim/.style={draw, fill=white, minimum width=6.5mm, minimum height=5.5mm, inner sep=1.5pt, font=\small},
  pcnode/.style={draw=pcline, fill=pcfill, diamond, aspect=1.6, inner sep=0.6pt, font=\scriptsize},
  marked/.style={draw=relline, line width=1.2pt},
  dead/.style={draw=deadgray, text=deadgray},
  ord/.style={-{Stealth[length=1.8mm]}, semithick},
  pcord/.style={-{Stealth[length=1.6mm]}, pcline, densely dashed},
  panel/.style={draw=black!35, rounded corners=4pt, inner sep=4pt},
  paneltitle/.style={font=\footnotesize, anchor=south west, inner sep=1pt},
  note/.style={font=\scriptsize, text=black!70},
}

\newtheorem{theorem}{Theorem}
\newtheorem{proposition}{Proposition}
\newtheorem{lemma}{Lemma}
\newtheorem{corollary}{Corollary}
\theoremstyle{definition}
\newtheorem{definition}{Definition}
\newtheorem{example}{Example}
\theoremstyle{remark}

\newcommand{\pre}{\mathit{pre}}
\newcommand{\add}{\mathit{add}}
\newcommand{\del}{\mathit{del}}
\newcommand{\pc}{\mathit{pc}}
\newcommand{\free}{\mathsf{free}}
\newcommand{\lock}{\mathsf{lock}}
\newcommand{\first}{\mathit{first}}
\newcommand{\Lang}{\mathcal{L}}
\newcommand{\Comp}{\mathcal{C}}
\newcommand{\Lfirst}{\Lang_{\mathrm{first}}}
\newcommand{\Lhddl}{\Lang_{\mathrm{HDDL}}}

\newenvironment{proofidea}{\par\medskip\noindent\textit{Proof idea.}\ }{\par\medskip}
\newenvironment{restate}[1]{\par\medskip\noindent\textbf{#1.}\ \itshape}{\par\smallskip}
\newcommand{\leanurl}{\url{https://github.com/HarrisonOates/method-preconditions-lean}}

\title{But It Was True When I Checked: Compiling SHOP-Style Method Preconditions into Partially-Ordered HTN Planning}
\author{Harrison Oates}
\affiliations{
    The Australian National University\\
    harrison.oates@anu.edu.au
}

\begin{document}

\maketitle

\begin{abstract}
Method preconditions specify when a method may be applied in hierarchical task network (HTN) planning, but their meaning differs under two established semantics.
The SHOP planners, like the classical HTN formalism of Erol, Hendler and Nau, require the precondition to hold immediately before the method's first action.
HDDL, the input language of the International Planning Competition's HTN track, instead introduces an auxiliary action to check it; under partial order, unrelated actions may invalidate the precondition before the method starts.
HDDL's designers adopted this weaker semantics because the first-action reading could not easily be compiled.
Subsequent work has repeatedly identified its compilation under partial order as an open question, with existing transformations limited to total order.
We give a compilation of the first-action semantics into standard partially ordered HTN planning without method preconditions using only two additional facts, thereby resolving a longstanding question in the HTN planning community.

\end{abstract}

\section{Introduction}\label{sec:intro}

Method preconditions restrict when a method may be used to decompose a task.
SHOP2 \citep{NauAIKMWY03} and SHOP3 \citep{GoldmanK19} rely on them heavily, and HDDL, the Hierarchical Domain Definition Language \citep{HoellerBBBPA20}, supports them through \texttt{:method-preconditions}.
HDDL defines their meaning by compilation: each method receives a new primitive task that carries the precondition and is ordered before all other subtasks of the method \citep{HoellerBBBPA20}.
Under partial order, this differs from SHOP2's semantics.

Under total order, the two readings coincide: only the auxiliary tasks of nested methods can occur between a method's check and its first action, and since these do not change the state, the precondition still holds when the first action executes.
Under partial order, however, unrelated actions may intervene between the check and the method's first action, as the following example shows.

\begin{example}[Shipping]\label{ex:ship}
A robot must ship an order and, independently, carry out maintenance on its loading bay; the two tasks are unordered.
Maintenance closes the bay and later reopens it.
Shipping has a single method that loads and dispatches the order, with the precondition that the bay is open.
Under the HDDL compilation, the plan \emph{check that the bay is open, close the bay, load, dispatch, reopen the bay} is a solution: the precondition is checked at the start, where it holds, and the order is then shipped through the closed bay.
The precondition held when it was checked, but not when the method's first action executed.
\end{example}

SHOP2 behaves differently.
It evaluates a method's precondition when it decomposes a task, and then restricts its next choice to the subtasks of that method \citep[Fig.~5]{NauAIKMWY03}.
Since decomposition does not change the state, the next action, counting the no-ops that represent empty methods, descends from that method and executes in the state in which its precondition was evaluated.
SHOP3 \citep{GoldmanK19} inherits this behaviour from SHOP2, and the original SHOP planner, which supports only totally ordered methods \citep{NauCLM99}, evaluates each precondition directly before the method's first action by construction.
We call this the \emph{first-action semantics}: a method's precondition must hold directly before the first action that the method produces.
For an empty method, we follow SHOP2 in evaluating the precondition when the task is removed.
The first-action reading already appears in the classical formalism of \citet{ErolHN94tr}, where a state constraint $(l,n)$ requires precondition $l$ to hold immediately before the first primitive task into which the abstract task $n$ is decomposed.
Such constraints can therefore express method preconditions.
The parsing-based verifier of \citet{BartakOMBB20} uses the same reading.

\paragraph{An open problem.}
Whether the first-action semantics can be compiled into standard HTN planning under partial order has remained open since HDDL was designed.
Its authors contrast the two readings and note that a system using the first-action reading ``needs to (natively) support this feature, because a compilation is not easily possible''; HDDL therefore adopts the compilation semantics and leaves the first-action reading as ``a feature for future extensions'' \citep[p.~9887]{HoellerBBBPA20}.
The rules of the IPC 2020 HTN track, whose input language IPC 2023 retained, accordingly give two different semantics: under total order the precondition must hold in the state in which the method's first action executes, but under partial order only in some state before it \citep{BehnkeHB21ipc}.
The problem has been raised repeatedly since.
\citet{HoellerWBB22} describe the semantics of method preconditions under partial order as ``a bit problematic'' and explain that HDDL's definition was chosen ``due to practical reasons, since it is the most simple way for the planning systems''.
\citet{ChenB21} observe that the HDDL reading compiles away easily, ``whereas the second version has no obvious compilation even in the deterministic setting''.
\citet{YousefiB24} exclude method preconditions from their formalism, describing the HDDL interpretation as ``semantically doubtful in a partial order setting (while being perfectly fine in total order)''.

Work on plan verification evaluates the first-action reading directly, but its extensions and transformations are likewise restricted to total order.
\citet{LinBOBB23} extend CYK-based verification to method preconditions under total order only.
\citet{OndrckovaB23} contrast decomposition constraints checked immediately before a task with HDDL's ``sometime before'' conditions, and give their transformations between constraint types for total order only.
\citet{OndrckovaPB24} show that interleaving makes the related compilation of constrained empty methods ``significantly more difficult'' under partial order and leave it open.%

Meanwhile, much of the theory and many planners for HTN planning are formulated without method preconditions, on the grounds that HDDL's auxiliary task compiles them away \citep{AlfordBHBBA16,BehnkeHB17,HoellerBBB18}; \citet{BehnkeHB17} call this ``not a real restriction''.
The argument is sound under total order, but under partial order it silently replaces the first-action semantics by HDDL's.
Consequently, the same HDDL file can denote different solution languages for a SHOP-style planner and a planner that follows the standard, with the latter accepting plans such as the one in \cref{ex:ship}.
Results stated for formalisms without method preconditions therefore do not directly carry over to the first-action semantics.

Any compilation of the first-action semantics must overcome three challenges.
First, a method's first action may lie several decompositions below one of several unordered subtasks, and which subtask produces it is only fixed by the plan.
Second, nested methods may share that first action, so their preconditions must hold in the same state.
Third, interleaving lets a task elsewhere in the hierarchy produce an action between a method's check and its first action, the obstacle identified by \citet{OndrckovaPB24}, and the compilation must exclude every such intervention, however many methods the plan applies.

We show that the first-action semantics can nevertheless be compiled exactly into standard partially ordered HTN planning without method preconditions, making it available to any planner for that formalism.
We address the first two challenges by letting each method nondeterministically \emph{mark} the subtask that will produce its first action and pass this obligation down the hierarchy.
A global lock, represented by just two fresh facts, addresses the third by protecting the interval from each precondition check to the marked action.
The task network's ordering constraints ensure that only the chain of method applications that acquires the lock can release it, so the lock need not record an owner.

\paragraph{Contributions.}
We formalise the first-action semantics, show that it refines HDDL's semantics and characterises SHOP2's runs (\cref{sec:semantics}), and give an exact compilation (\cref{sec:compilation}) with a bijection between solutions and a machine-checked proof of language preservation (\cref{sec:correctness}).
The compilation has polynomial size and stays within the totally ordered, acyclic, tail-recursive and regular fragments, so known plan-existence bounds carry over to the first-action semantics (\cref{sec:size}).
It also applies unchanged to lifted HDDL domains, so it can be run on HDDL domain files without grounding them first.
Proof ideas are given in the main text and full proofs in \crefrange{app:semantics}{app:size}.

\section{Preliminaries}\label{sec:prelims}

We use the ground HTN formalism of \citet{GeierB11}, with an initial task network, and extend methods with positive conjunctive preconditions.

\paragraph{Problems and actions.}
A problem is a tuple $P = (F, N_C, N_P, \delta, M, s_I, tn_I)$, where $F$ is a finite set of facts, $N_C$ and $N_P$ are disjoint finite sets of compound and primitive task names, $M$ is a finite set of methods, $s_I \subseteq F$ is the initial state, and $tn_I$ is the initial task network.
Each primitive task name $o \in N_P$ denotes an action with description $\delta(o) = (\pre(o), \add(o), \del(o))$, where each component is a subset of $F$.
The action is applicable in a state $s \subseteq F$ if $\pre(o) \subseteq s$, and it then leads to the state $(s \setminus \del(o)) \cup \add(o)$.
We identify primitive task names with their actions.

\paragraph{Task networks.}
A \emph{task network} $tn = (T, \prec, \alpha)$ consists of a set $T$ of task identifiers, a strict partial order $\prec$ on $T$, and a labelling $\alpha : T \to N_C \cup N_P$.
Identifiers distinguish task occurrences, allowing two tasks with the same name to have different positions in the network.
We write $\mathrm{Min}(T,\prec)$ for the \emph{minimal} tasks, those with no predecessor in $T$.
A \emph{linearisation} of $tn$ is a sequence that contains every identifier exactly once and respects $\prec$.

\paragraph{Methods and decomposition.}
A method $m = (c, \pre(m), tn_m)$ consists of a compound task name $c$, its \emph{head}, a precondition $\pre(m) \subseteq F$, and a subtask network $tn_m = (T_m,\prec_m,\alpha_m)$.
A task $t$ labelled $c$ is \emph{decomposed} with $m$ by replacing $t$ with a fresh copy of $tn_m$.
The internal order of $tn_m$ is retained, every predecessor of $t$ is ordered before every new subtask, and every successor of $t$ is ordered after every new subtask.
We call each such step a \emph{method application}.
Decomposition itself ignores the state and $\pre(m)$; the two semantics in \cref{sec:semantics} determine where the precondition must hold.

A \emph{complete decomposition} is a finite sequence of method applications, starting from $tn_I$ and ending in a network whose tasks are all primitive.
We keep track of the decomposition forest, so every task occurrence has a unique chain of ancestors.
The \emph{yield} of a method application $\mu$ is the set of primitive occurrences that descend from its subtasks.
We write $\pre(\mu)$ for the precondition of the method used by $\mu$.

The proofs repeatedly use the following property of inherited orderings.
If two tasks $u \prec v$ occur in some network during the decomposition, then in the final network every primitive descendant of $u$ precedes every primitive descendant of $v$; the same holds when $u$ or $v$ is itself primitive.
This follows by induction, because decomposition copies the orderings of a task to all of its subtasks.

\paragraph{Solutions.}
A linearisation $a_1, \ldots, a_\ell$ of the final network is \emph{executable} if the actions can be applied in this order starting from $s_I$.
We then write $s_0 = s_I$ and $s_i$ for the state after $a_1,\ldots,a_i$.
For a problem without method preconditions, a \emph{solution} is a complete decomposition together with an executable linearisation of its final network, and the \emph{solution language} $\Lang(P)$ is the set of action words $\alpha(a_1)\cdots\alpha(a_\ell)$ of all solutions \citep{HoellerBBB14,HoellerBBB16}.

A problem is \emph{totally ordered} if its initial network and all method networks are totally ordered.
It is \emph{acyclic} if the graph on $N_C$ with an edge $c \to d$ whenever some method for $c$ has a subtask labelled $d \in N_C$ contains no directed cycle.
A subtask of a method is \emph{last} if every other subtask precedes it.
A problem is \emph{tail-recursive} if some level function on $N_C$ gives every compound subtask of a method for $c$ a lower level than $c$, except a last subtask, whose level may equal that of $c$ \citep{AlfordBA15}.
It is \emph{regular} if every method has at most one compound subtask, which is last \citep{ErolHN96}.

\paragraph{Empty methods.}
Since an empty method has no first action, we must specify where its precondition is evaluated.
We replace every empty method by one whose only subtask is a fresh no-op action $\mathit{nop} = (\emptyset,\emptyset,\emptyset)$, and erase no-ops only when reporting solution words.
This mirrors SHOP2, which evaluates an empty method's precondition when it removes the task \citep[Fig.~5]{NauAIKMWY03}, and the parsing-based verifiers, which place empty methods between two actions \citep{BartakOMBB20,OndrckovaB23}.
This convention is part of the semantics: the no-op determines the state of the check and can also be the first action of an ancestor.
For example, let $q$ hold initially, let $D$ have a single method with precondition $q$ and subtasks $C \prec r$, let $C$ have an empty method, and let an action $k$ unordered with $D$ delete $q$.
The word $k\,r$ is a solution because the no-op of $C$ can run before $k$, while $q$ still holds.
Requiring $q$ to hold before $r$, the first action that is not a no-op, would reject this word (\cref{sec:discussion}).
From now on, every method has at least one subtask, so every method application has a first action in any linearisation.

\paragraph{The HDDL compilation.}
For a problem $P$ with method preconditions, the HDDL compilation $H(P)$ replaces each method $m = (c, \pre(m), tn_m)$ with a method without precondition.
Its network extends $tn_m$ with one fresh task $t_m$ labelled by the auxiliary action $\pc_m = (\pre(m), \emptyset, \emptyset)$ and orders $t_m$ before every task of $tn_m$ \citep{HoellerBBBPA20}.
We call $\pc_m$ the \emph{check} of $m$.
Decompositions of $P$ and of $H(P)$ correspond one to one: they make the same method choices, and every application $\mu$ in $H(P)$ additionally contains a check task $t_\mu$.
The HDDL semantics is
\[
  \Lang_{\mathrm{HDDL}}(P) = h(\Lang(H(P))),
\]
where $h$ erases all checks.

\section{Two Readings of Method Preconditions}\label{sec:semantics}

We formalise the two readings using the shipping example and relate them to SHOP2's task-selection rule.

\subsection{A Running Example}

\begin{figure}[t]
\centering
\resizebox{\columnwidth}{!}{\begin{tikzpicture}
  \def\pw{3.9}\def\ph{1.5}
  \foreach \x/\y/\n in {0/0/pa, 4.4/0/pb, 0/-2.2/pc, 4.4/-2.2/pd} {
    \node[panel, minimum width=\pw cm, minimum height=\ph cm, anchor=south west, inner sep=0pt] (\n) at (\x,\y) {};
  }
  \node[paneltitle] at (pa.north west) {(a) initial network};
  \node[paneltitle] at (pb.north west) {(b) method $m_S$ of $S$};
  \node[paneltitle] at (pc.north west) {(c) method $m_U$ of $U$};
  \node[paneltitle] at (pd.north west) {(d) method $m_M$ of $M$};
  \node[note, anchor=north east] at (pa.north east) {$s_I = \{p\}$};
  \node[note, anchor=north east] at (pb.north east) {$\mathit{pre} = \{p\}$};
  \node[note, anchor=north east] at (pc.north east) {$\mathit{pre} = \{p\}$};
  \node[note, anchor=north east] at (pd.north east) {$\mathit{pre} = \emptyset$};

  \node[comp] at ($(pa.center)+(-0.6,-0.12)$) {$S$};
  \node[comp] at ($(pa.center)+(0.6,-0.12)$) {$M$};
  \node[comp] at ($(pb.center)+(-0.6,-0.12)$) {$U$};
  \node[prim] at ($(pb.center)+(0.6,-0.12)$) {$c$};
  \node[prim] (a) at ($(pc.center)+(-0.6,-0.12)$) {$a$};
  \node[prim] (b) at ($(pc.center)+(0.6,-0.12)$) {$b$};
  \draw[ord] (a) -- (b);
  \node[prim] (k) at ($(pd.center)+(-0.6,0.0)$) {$k$};
  \node[prim] (r) at ($(pd.center)+(0.6,0.0)$) {$r$};
  \draw[ord] (k) -- (r);
  \node[note, below=0.3mm of k] {deletes $p$};
  \node[note, below=0.3mm of r] {adds $p$};
\end{tikzpicture}}
\caption{The running example.
Rounded boxes are compound tasks, square boxes are actions, and arrows are ordering constraints.
$S$ and $M$ are unordered, and so are $U$ and $c$ in the method of $S$.
The actions $a$, $b$, $c$ have no preconditions or effects.}
\label{fig:example}
\end{figure}
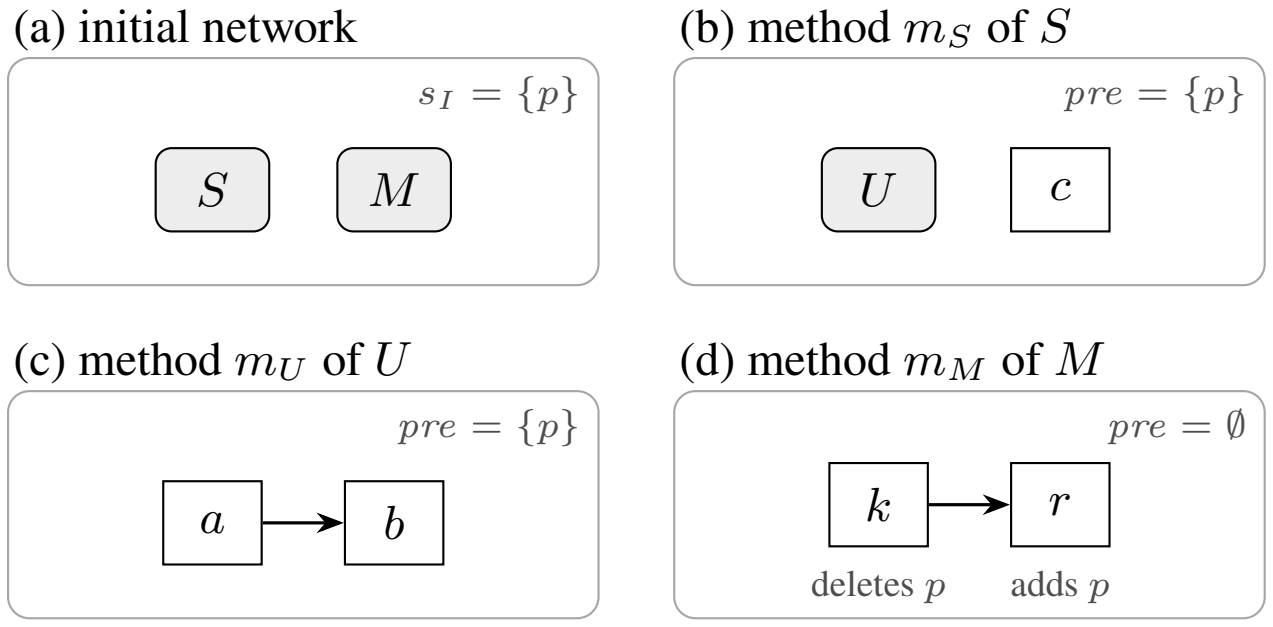

\Cref{fig:example} formalises the shipping scenario of \cref{ex:ship}.
There is a single fact $p$, ``the loading bay is open'', which is initially true.
The initial network contains two unordered compound tasks: the shipment $S$ and the maintenance $M$.
Maintenance has a single method $m_M$ without precondition, which executes $k$, deleting $p$, and then $r$, adding $p$ again.
The shipment has a single method $m_S$ with precondition $p$ and two unordered subtasks: a compound task $U$ and an action $c$.
The task $U$ has a single method $m_U$, also with precondition $p$, whose subtasks are the actions $a \prec b$.
No action has a precondition, so every order-respecting interleaving is executable; only the method preconditions can exclude it as a solution.

There is just one complete decomposition.
Its final network consists of the chains $a \prec b$ and $k \prec r$ together with the action $c$, which is unordered with all of them.
These five actions have $5!/(2!\cdot 2!) = 30$ linearisations.

\subsection{The HDDL Semantics}

In $H(P)$, the methods $m_S$ and $m_U$ receive checks $\pc_S$ and $\pc_U$ that test $p$.
The check $\pc_S$ must precede $U$ and $c$, and hence all of $a$, $b$, $c$ and $\pc_U$; the check $\pc_U$ must precede $a$ and $b$.
Neither check is ordered with respect to $k$ or $r$.
Both checks can therefore be scheduled at the beginning, where $p$ holds, making all $30$ linearisations solutions under the HDDL semantics.
\Cref{fig:semantics}(i) shows one of them: the shipment starts with $a$ while the bay is closed.

\subsection{The First-Action Semantics}

The first-action semantics requires each method's precondition to hold in the state directly before the method's first action.

\begin{definition}[First-action semantics]\label{def:imm}
Consider a complete decomposition of $P$ and an executable linearisation $a_1,\ldots,a_\ell$ of its final network.
For a method application $\mu$, let
\[
  \first(\mu) = \min\{\, i \mid a_i \text{ is in the yield of } \mu \,\}
\]
be the position of its first action.
The pair is a \emph{first-action solution} if $\pre(\mu) \subseteq s_{\first(\mu)-1}$ for every method application $\mu$.
The \emph{first-action language} $\Lfirst(P)$ is the set of action words of all first-action solutions, with no-ops erased.
\end{definition}

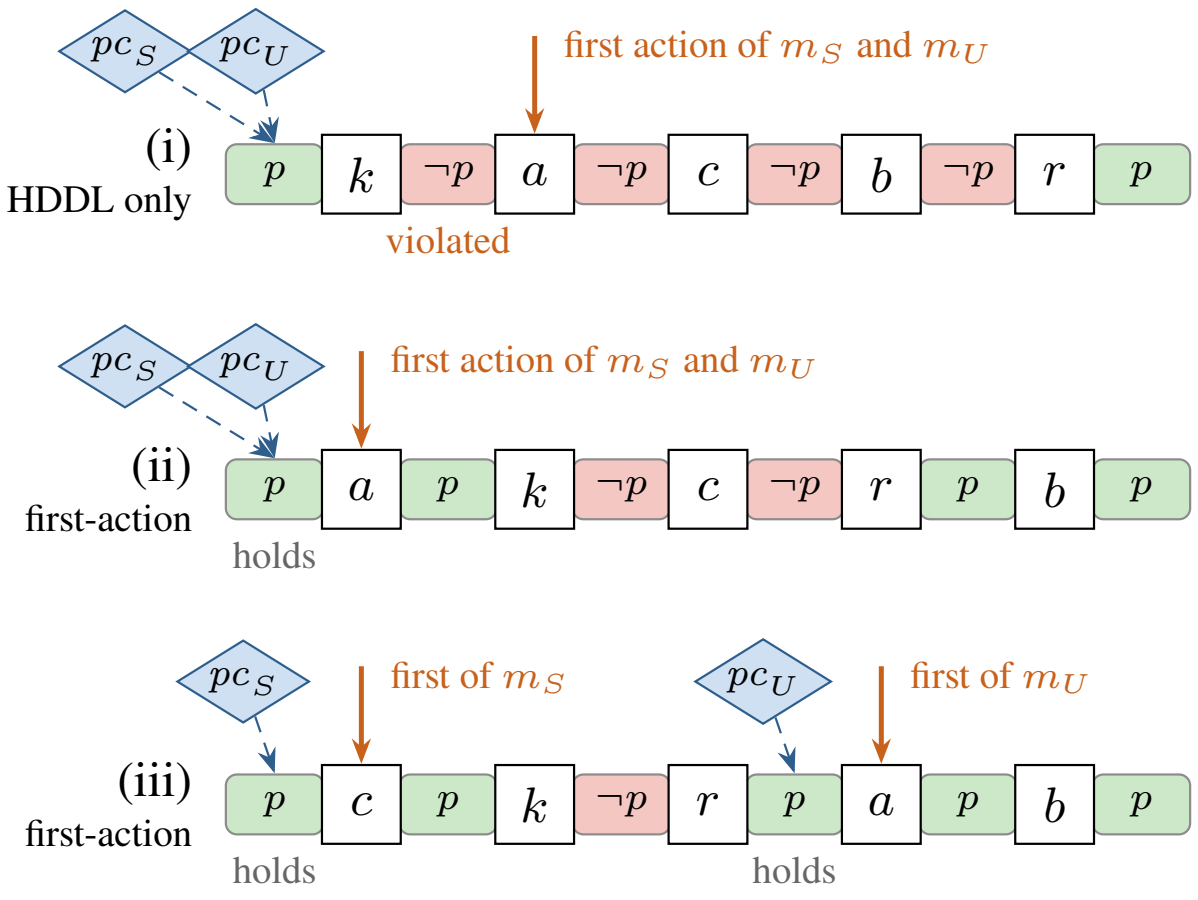
\begin{figure}[t]
\centering
\resizebox{\columnwidth}{!}{\def\TT{T}
\newcommand{\semrow}[4]{%
  \node[font=\footnotesize, anchor=east, align=right] at (-0.4,#1) {#2};
  \foreach \s [count=\i from 0] in {#4} {
    \pgfmathsetmacro{\xx}{1.1*\i}
    \edef\st{\s}
    \ifx\st\TT
      \node[draw=black!45, fill=ptrue, rounded corners=2pt, minimum width=6.2mm, minimum height=3.8mm, inner sep=0pt, font=\scriptsize]
        (st\i) at (\xx,#1) {$p$};
    \else
      \node[draw=black!45, fill=pfalse, rounded corners=2pt, minimum width=6.2mm, minimum height=3.8mm, inner sep=0pt, font=\scriptsize]
        (st\i) at (\xx,#1) {$\neg p$};
    \fi
  }
  \foreach \a [count=\i from 1] in {#3} {
    \pgfmathsetmacro{\xx}{1.1*\i-0.55}
    \node[prim, minimum width=5mm, minimum height=5mm] (ac\i) at (\xx,#1) {$\a$};
  }
}
\newcommand{\pcat}[3]{%
  \node[pcnode] (pcm) at ($(st#1)+(#2,0.78)$) {#3};
  \draw[pcord] (pcm) -- (st#1.north);
}
\newcommand{\firstof}[2]{%
  \draw[relline, line width=0.9pt, -{Stealth[length=1.6mm]}] ($(ac#1.north)+(0,0.62)$) -- (ac#1.north);
  \node[font=\scriptsize, text=relline, anchor=west] at ($(ac#1.north)+(0.06,0.55)$) {#2};
}
\begin{tikzpicture}
  \semrow{0}{(i)\\[-1pt]\scriptsize HDDL only}{k,a,c,b,r}{T,F,F,F,F,T}
  \pcat{0}{-0.95}{$\mathit{pc}_S$}
  \pcat{0}{-0.12}{$\mathit{pc}_U$}
  \firstof{2}{first action of $m_S$ and $m_U$}
  \node[font=\scriptsize, text=relline] at ($(st1.south)+(0,-0.22)$) {violated};

  \semrow{-2.0}{(ii)\\[-1pt]\scriptsize first-action}{a,k,c,r,b}{T,T,F,F,T,T}
  \pcat{0}{-0.95}{$\mathit{pc}_S$}
  \pcat{0}{-0.12}{$\mathit{pc}_U$}
  \firstof{1}{first action of $m_S$ and $m_U$}
  \node[font=\scriptsize, text=black!60] at ($(st0.south)+(0,-0.22)$) {holds};

  \semrow{-4.0}{(iii)\\[-1pt]\scriptsize first-action}{c,k,r,a,b}{T,T,F,T,T,T}
  \pcat{0}{-0.2}{$\mathit{pc}_S$}
  \firstof{1}{first of $m_S$}
  \pcat{3}{-0.2}{$\mathit{pc}_U$}
  \firstof{4}{first of $m_U$}
  \node[font=\scriptsize, text=black!60] at ($(st0.south)+(0,-0.22)$) {holds};
  \node[font=\scriptsize, text=black!60] at ($(st3.south)+(0,-0.22)$) {holds};
\end{tikzpicture}}
\caption{Three linearisations of the example.
The small boxes between the actions are the states, green where $p$ holds and red where it does not.
Diamonds are checks, drawn at the state in which they are evaluated.
(i) is a solution under the HDDL semantics but not under the first-action one; (ii) and (iii) are solutions under both.
The vacuous check of $m_M$ is omitted.
The labels $p$ and $\neg p$ distinguish the states without colour.}
\label{fig:semantics}
\end{figure}

In the example, the first action of $m_U$ is always $a$, and the first action of $m_S$ is whichever of $a$ and $c$ comes first.
Thus $a$ cannot execute strictly between $k$ and $r$, where $p$ is false, and neither can $c$ if it precedes $a$.
This rules out $11$ of the $30$ linearisations and leaves $19$.
In linearisation (i) of \cref{fig:semantics}, $a$ is the first action of both $m_S$ and $m_U$ and comes after $k$, where $p$ is false; HDDL accepts this plan because both checks may run before $k$.
In (ii), both methods start with $a$, so both preconditions must hold in the initial state.
In (iii), $m_S$ starts with $c$, while $m_U$ starts with $a$ after $r$; their preconditions are therefore evaluated in different states.

\subsection{Comparing the Two Semantics}

To compare the two semantics, we insert the checks of $H(P)$ into a linearisation of $P$, each directly before its method's first action.
This placement respects the ordering constraints of $H(P)$, and we use it again in \cref{sec:correctness}.

\begin{lemma}[Check insertion]\label{lem:insert}
Take a complete decomposition of $P$, a linearisation $a_1,\ldots,a_\ell$ of its final network, and the corresponding decomposition of $H(P)$.
Placing each check $t_\mu$ directly before $a_{\first(\mu)}$, with the checks placed before the same action ordered from ancestor to descendant, gives a linearisation of the final network of $H(P)$.
\end{lemma}

\begin{proposition}\label{prop:compare}
$\Lfirst(P) \subseteq \Lhddl(P)$.
The two languages coincide if $P$ is totally ordered, and the inclusion can be strict otherwise.
\end{proposition}

\begin{proofidea}
For the inclusion, insert the checks as in \cref{lem:insert}; each check $t_\mu$ is then evaluated in $s_{\first(\mu)-1}$, which satisfies $\pre(\mu)$.
Under total order, every occurrence outside $\mu$'s subtree is ordered before $t_\mu$ or after the whole yield.
Hence only descendant checks can occur between $t_\mu$ and the first action of $\mu$, and these leave the state unchanged.
Linearisation (i) of the running example shows strictness.
\end{proofidea}

\subsection{Relation to SHOP2}\label{sec:shop}

SHOP2 maintains a set of tasks that it may work on next, initially the minimal tasks of the network \citep[Fig.~5]{NauAIKMWY03}.
When it decomposes a compound task, it evaluates the method's precondition in the current state and restricts its next choice to the minimal subtasks of that method.
When it executes a primitive task, or removes a task with an empty method, it again allows any minimal remaining task.
We apply this rule to arbitrary partial orders and call a sequence of such steps that ends with an empty network a \emph{run}.

\begin{proposition}\label{prop:shop}
For a ground problem $P$, the action words of the runs of SHOP2's task-selection rule are exactly $\Lfirst(P)$.
\end{proposition}

\begin{proofidea}
After a decomposition, SHOP2 continues below that method until it executes a primitive task, without changing the state along the way.
That action is therefore the first action of every method on the path, and executes in the state in which their preconditions were evaluated.
Conversely, replay a first-action solution and, before each $a_i$, decompose from the top down all tasks whose method applications have $a_i$ as their first action.
These applications lie consecutively on the ancestor path of $a_i$, and each chosen task is minimal, so the resulting sequence is a run.
Removing a task with an empty method corresponds to executing its no-op.
\end{proofidea}

Goldman's formal semantics of SHOP2 makes the same choice: it treats the precondition test as atomic with the method's expansion, so that the precondition cannot be clobbered before the method's first action \citep{Goldman09}.

\section{The Compilation}\label{sec:compilation}

We now compile the first-action semantics into a problem $\Comp_{\pc}(P)$ without method preconditions.
As in the HDDL compilation, each precondition is tested by a check action ordered before the method's subtasks.
To enforce the first-action semantics, we also prevent unrelated actions from executing between this check and the method's first action.

\subsection{Selecting the First Subtask}

The subtask that produces a method's first action depends on the plan: for $m_S$, it is $U$ in some plans and $c$ in others.
We represent this choice by replacing each method with several \emph{variants}, one for each subtask $t^*$ that might produce the first action, and call the selected subtask \emph{marked}.
Only minimal subtasks need to be considered, since a subtask with a predecessor $u$ comes after the entire non-empty yield of $u$ and cannot contain the first action.

If the marked subtask is primitive, it is the method's first action.
If it is compound, the obligation passes to the method that decomposes it, whose first action must also be the first action of the parent method.
To record this, the compiled problem has two copies of every compound task name $c$:
\begin{itemize}[nosep]
\item the \emph{normal} copy $c^N$, used when $c$ carries no obligation, and
\item the \emph{first} copy $c^F$, used when the first action produced by $c$ must also be the first action of an ancestor whose check has already been executed.
\end{itemize}
In each variant, the marked subtask becomes $d^F$ if it is a compound task $d$, and $o^R$ if it is an action $o$; the superscript $R$ stands for \emph{release}, for reasons explained below.
All other subtasks become normal copies.

Following the marks downwards from a normal task gives a path through zero or more first copies that ends at a release action.
We call it a \emph{marked chain}.
In the example, marking $U$ in $S^N$ gives the chain $m_S,m_U,a^R$, as in linearisation (ii).
Marking $c$ instead gives $m_S,c^R$ and a separate chain $m_U,a^R$, as in (iii).

\subsection{Protecting the Interval before the First Action}

Marking designates the intended first action, but does not by itself enforce that choice.
We enforce it with two fresh facts, $\free$ and $\lock$, exactly one of which holds at any time.
\begin{itemize}[nosep]
\item The \emph{opening} check at the start of a chain, in a method for a normal task, requires the method's precondition and $\free$, and replaces $\free$ by $\lock$.
\item An \emph{inner} check, in a method for a first copy, requires the precondition and $\lock$, and changes nothing.
\item A release action $o^R$ requires $\lock$ in addition to the precondition of $o$, and replaces $\lock$ by $\free$.
\item Every other action $o^N$ requires $\free$ in addition to the precondition of $o$.
\end{itemize}
We call the interval from the opening check to the release a \emph{lock window}.
Inside a window, normal actions and opening checks are blocked, so only inner checks and releases can run.

The task network makes an explicit lock owner unnecessary.
Every other chain is either unopened, in which case its opening check prevents its inner checks and release from running, or completed, in which case they have already run.
Only the active chain can therefore release the lock, and two facts suffice regardless of the number of method applications (\cref{lem:window}).

This restriction applies only within lock windows: the compilation adds no ordering between the marked subtask and its siblings, so the remaining yield can still interleave with other tasks.

\subsection{Formal Construction}

\begin{definition}[$\Comp_{\pc}(P)$]\label{def:comp}
Let $P$ be a problem in which every method has at least one subtask, and let $\free$, $\lock$ and all task names introduced below be fresh.
The compiled problem $\Comp_{\pc}(P)$ has facts $F' = F \cup \{\free,\lock\}$, initial state $s_I' = s_I \cup \{\free\}$, and compound task names
\[
  N_C' = \{c^N,c^F \mid c \in N_C\}.
\]
Its primitive task names are the following, where each triple lists precondition, add effects and delete effects.
For every action $o \in N_P$, there are a normal copy and a release copy,
\begin{align*}
  o^N &= (\pre(o) \cup \{\free\},\;\add(o),\;\del(o)),\\
  o^R &= (\pre(o) \cup \{\lock\},\;\add(o) \cup \{\free\},\\
  &\phantom{{}= (}\del(o) \cup \{\lock\}),
\end{align*}
and for every method $m \in M$ there are an opening check and an inner check,
\begin{align*}
  \pc_m^N &= (\pre(m) \cup \{\free\},\;\{\lock\},\;\{\free\}),\\
  \pc_m^F &= (\pre(m) \cup \{\lock\},\;\emptyset,\;\emptyset).
\end{align*}

For every method $m=(c,\pre(m),(T,\prec,\alpha))$, every mode $X\in\{N,F\}$, and every marked task $t^*\in\mathrm{Min}(T,\prec)$, there is a method $m_{X,t^*}$ for $c^X$ without precondition.
Its network $(T',\prec',\alpha')$ adds a fresh check task $t_m$ before all subtasks and relabels the subtasks according to the marking:
\begin{align*}
  T' &= T\cup\{t_m\},\\
  {\prec'} &= {\prec}\;\cup\;\{(t_m,t)\mid t\in T\},\\
  \alpha'(t) &=
  \begin{cases}
    \pc_m^X & t=t_m,\\
    \alpha(t)^F & t=t^* \text{ and } \alpha(t)\in N_C,\\
    \alpha(t)^R & t=t^* \text{ and } \alpha(t)\in N_P,\\
    \alpha(t)^N & t\in T\setminus\{t^*\}.
  \end{cases}
\end{align*}
The initial network keeps the task identifiers and order of $tn_I$ and replaces each label $x$ by $x^N$.

The projection $h$ erases all checks and maps $o^N$ and $o^R$ to $o$ for every $o \in N_P$, erasing the copies of $\mathit{nop}$.
\end{definition}

Marks select subtask \emph{occurrences}; a variant whose marked subtask cannot supply an applicable first action has no executable completion.
A window opens only when a check $\pc_m^N$ is executed, not when $c^N$ is decomposed.
Since the compilation copies $\pre(m)$ to a check without inspecting its structure, it applies unchanged to any precondition language supported by the target planner's actions, including negation, disjunction and quantification in HDDL.

\subsection{The Compiled Example}

\begin{figure}[t]
\centering
\begin{tikzpicture}[
  lab/.style={font=\small, anchor=east},
  cmp/.style={comp, minimum width=8mm},
  prm/.style={prim, minimum width=7mm},
]
  \node[font=\footnotesize\bfseries, anchor=west] at (-1.0,0.85) {methods of $S^N$};
  \node[lab] at (0,0) {$S^N \mapsto$};
  \node[pcnode] (p1) at (0.55,0) {$\mathit{pc}^N_S$};
  \node[cmp, marked] (u1) at (1.75,0.33) {$U^F$};
  \node[prm] (c1) at (1.75,-0.33) {$c^N$};
  \draw[pcord] (p1) -- (u1.west); \draw[pcord] (p1) -- (c1.west);
  \node[lab] at (0,-1.25) {$S^N \mapsto$};
  \node[pcnode] (p2) at (0.55,-1.25) {$\mathit{pc}^N_S$};
  \node[cmp] (u2) at (1.75,-0.92) {$U^N$};
  \node[prm, marked] (c2) at (1.75,-1.58) {$c^R$};
  \draw[pcord] (p2) -- (u2.west); \draw[pcord] (p2) -- (c2.west);

  \begin{scope}[xshift=3.45cm]
  \node[font=\footnotesize\bfseries, anchor=west] at (-1.0,0.85) {methods of $U^N$, $U^F$, $M^N$};
  \foreach \X/\Y/\pcl in {N/0.1/{\mathit{pc}^N_U}, F/-0.75/{\mathit{pc}^F_U}} {
    \node[lab] at (0,\Y) {$U^\X \mapsto$};
    \node[pcnode] (q) at (0.55,\Y) {$\pcl$};
    \node[prm, marked] (a) at (1.65,\Y) {$a^R$};
    \node[prm] (b) at (2.7,\Y) {$b^N$};
    \draw[pcord] (q) -- (a); \draw[ord] (a) -- (b);
  }
  \node[lab] at (0,-1.6) {$M^N \mapsto$};
  \node[prm] (k3) at (0.55,-1.6) {$k^N$};
  \node[prm] (r3) at (1.6,-1.6) {$r^N$};
  \draw[ord] (k3) -- (r3);
  \end{scope}
\end{tikzpicture}
\caption{Reachable methods of the compiled example.
Diamonds are checks, dashed arrows their ordering before the other subtasks, and orange boxes mark the subtask that has to produce the first action and release the lock.
The method of $M^N$ has no check because $m_M$ has no precondition.}
\label{fig:compiled}
\end{figure}
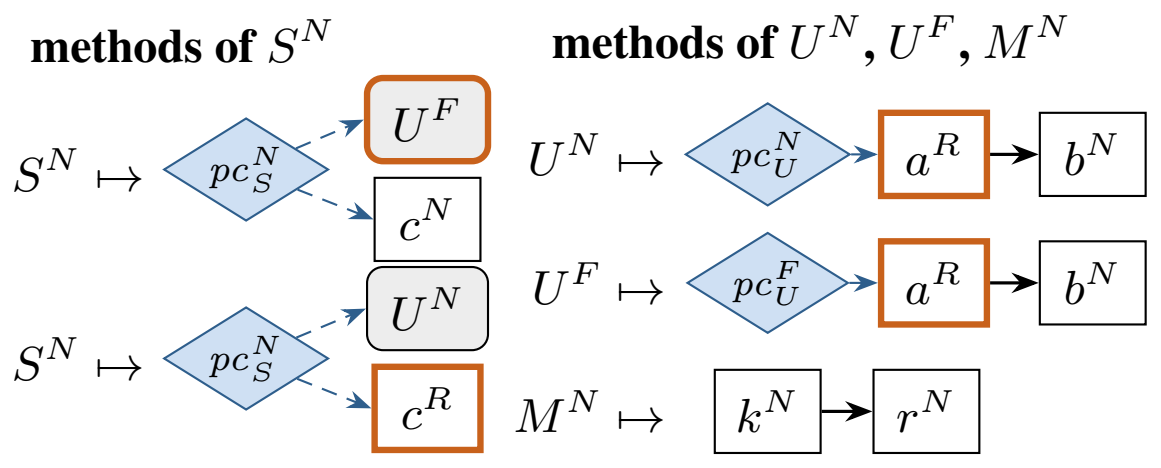

\Cref{fig:compiled} shows the reachable methods, omitting the vacuous check of $m_M$ as in the optimisation of \cref{prop:trivial}.
The task $S^N$ has the two variants discussed above, marking $U$ or $c$.
The tasks $U^N$ and $U^F$ have one variant each, which marks $a$, because $b$ is not minimal.
The copies $S^F$ and $M^F$ are unreachable, since $S$ and $M$ occur only in the initial network.
In total, the compiled problem has $8$ methods.

Linearisation (ii) of \cref{fig:semantics} uses the variant of $S^N$ that marks $U$ and the variant of $U^F$ that marks $a$: the check $\pc^N_S$ opens a window, $\pc^F_U$ checks $p$ inside it, and $a^R$ closes it, so both preconditions are evaluated directly before $a$.
Linearisation (iii) uses the variant of $S^N$ that marks $c$ and has two windows, $\pc^N_S\,c^R$ and, after $r$ has restored $p$, $\pc^N_U\,a^R$.
Linearisation (i) cannot be reproduced.
If $\pc^N_S$ opens a window before $k$, then $k^N$ must wait for $\free$ to be restored by the release, which is the first action of $m_S$.
If the check is delayed until after $k$, it fails while $p$ remains false.
Exhaustive enumeration confirms that the compiled solutions project to exactly the $19$ first-action solutions, each with a single compiled decomposition.

\section{Correctness}\label{sec:correctness}

The task network orders each marked chain (\cref{lem:chains}), and the lock makes each window contain exactly that chain (\cref{lem:window}), giving soundness.
For completeness, we construct the marking and windows of any first-action solution (\cref{thm:main}).
Full proofs are in \cref{app:correctness}.

Throughout this section, chains, checks and actions refer to occurrences in a fixed complete decomposition.
Decompositions of $P$ and of $\Comp_{\pc}(P)$ are related as follows.
Every compiled decomposition has an \emph{underlying} decomposition of $P$, obtained by erasing the check tasks and the annotations $N$, $F$ and $R$.
Conversely, a decomposition of $P$ together with a choice of one minimal marked subtask for every method application determines a compiled decomposition: label the initial tasks $N$, and then label the subtasks of each application as in \cref{def:comp}, from the top down.
Both translations preserve task identifiers and all ordering constraints between them.

\begin{lemma}[Marked chains]\label{lem:chains}
In a complete decomposition of $\Comp_{\pc}(P)$, each application to a task in mode $N$ begins a unique marked chain.
The chain passes through zero or more applications to tasks in mode $F$ and ends at one action occurrence in mode $R$.
If the chain's method applications are $\mu_0,\ldots,\mu_k$ and its release is $r$, then
\begin{equation}\label{eq:chain-order}
  \pc_{\mu_0}^N \prec \pc_{\mu_1}^F \prec \cdots
  \prec \pc_{\mu_k}^F \prec r
\end{equation}
in the final network.
Every check occurrence and every release occurrence belongs to exactly one marked chain.
\end{lemma}

\begin{proofidea}
Marks form a forest of downward paths: every application has exactly one marked child, only marked children are in mode $F$ or $R$, and the initial tasks are in mode $N$.
The ordering follows because every check precedes its method's subtasks and, by inherited ordering, all of their descendants.
\end{proofidea}

\begin{lemma}[Lock ownership]\label{lem:window}
In every executable linearisation of a complete decomposition of $\Comp_{\pc}(P)$, each lock window is exactly
\(
  \pc_{\mu_0}^N,\;\pc_{\mu_1}^F,\;\ldots,\;\pc_{\mu_k}^F,\;r
\)
for one marked chain of \cref{lem:chains}.
In particular, no window contains an action, check or release from outside its chain.
\end{lemma}

\begin{proofidea}
Induction along the linearisation shows that at most one chain is \emph{active}, meaning opened but not yet released, and that $\lock$ holds exactly when a chain is active.
While no chain is active, $\lock$ is false, so only normal actions and opening checks can run.
While a chain $C$ is active, $\free$ is false, so only inner checks and releases can run.
By \cref{eq:chain-order}, the inner checks and release of an unopened chain must wait for its opening check, while those of a completed chain have already run.
Hence every occurrence within the window belongs to $C$.
\end{proofidea}

\begin{theorem}[Soundness and completeness]\label{thm:main}
For every ground problem $P$,
\(
  h\bigl(\Lang(\Comp_{\pc}(P))\bigr)=\Lfirst(P).
\)
\end{theorem}

\begin{proofidea}
\emph{Soundness.}
Erasing checks and annotations gives a decomposition and linearisation of $P$ with the same effects on $F$.
For an application $\mu$, let $r$ be the release of its chain.
The release descends from $\mu$, whose check precedes every action in its yield.
By \cref{lem:window}, only checks occur between this check and $r$, so $r$ is the first action of $\mu$ and $\pre(\mu)$ still holds directly before it.

\emph{Completeness.}
Mark in every application $\mu$ the subtask containing $a_{\first(\mu)}$, which is minimal, and insert the checks by \cref{lem:insert}.
The applications whose first action is $a_i$ form a segment $\mu_0,\ldots,\mu_k$ of the ancestor path of $a_i$; $\mu_0$ is applied to a task in mode $N$, its descendants on the segment to tasks in mode $F$, and $a_i$ is in mode $R$.
So the compiled linearisation consists of normal actions and of blocks $\pc^N,\pc^F,\ldots,\pc^F,a_i^R$, each of which is executable from a state with $\free$ and restores it.
\end{proofidea}

Every chain releases the lock, so no goal condition on the lock is needed.
Moreover, the correspondence preserves individual solutions as well as their action words.

\begin{corollary}\label{cor:bijection}
Identify solutions by their decomposition forest and linearisation.
Then erasing checks and annotations is a bijection from the solutions of $\Comp_{\pc}(P)$ to the first-action solutions of $P$.
\end{corollary}

\begin{proofidea}
The soundness argument uniquely determines each application's mark as the subtask containing its first action, and the marks determine the modes.
By \cref{lem:window}, the checks form a block directly before their chain's release, in chain order, so their positions are also determined.
\end{proofidea}

\paragraph{Machine-checked proof.}
We formalised \cref{def:imm,def:comp}, \cref{lem:insert,lem:chains,lem:window}, \cref{prop:compare} and \cref{thm:main} in Lean~4 with Mathlib, the development is available at \leanurl.

\section{Size and Structure}\label{sec:size}

\begin{proposition}\label{prop:size}
Let $k_m=|\mathrm{Min}(T_m,\prec_m)|$ be the number of minimal subtasks of method $m$.
The compilation generates $2|N_C|$ compound task names, $2|N_P|+2|M|$ primitive task names and $2\sum_{m\in M}k_m$ methods.
With explicit task and ordering lists, its description size is
\begin{equation}\label{eq:size}
  O\!\left(|P|+\sum_{m\in M}k_m\bigl(|T_m|+|\prec_m|\bigr)\right).
\end{equation}
It can be constructed in polynomial time, has size $O(|P|^2)$ in general and $O(|P|)$ if every method is totally ordered, and preserves total order, acyclicity, tail recursion and regularity.
\end{proposition}

\begin{proofidea}
Each method has two variants per minimal subtask, each adding one task and $|T_m|$ ordering pairs.
Since $k_m \le |T_m|$, the size is quadratic in general; under total order, $k_m = 1$ gives a linear bound.
The added check is primitive and precedes every subtask, preserving total order and regularity and leaving a last subtask last.
Giving $c^N$ and $c^F$ the level of $c$ preserves tail recursion, while acyclicity follows because compiled dependencies project to original ones.
\end{proofidea}

\paragraph{Omitting vacuous checks.}
A method without a precondition needs no check of its own, but its $F$-variants must still pass on an ancestor's first-action obligation.

\begin{proposition}[Methods without preconditions]\label{prop:trivial}
For every method $m$ with $\pre(m)=\emptyset$, replace its $N$-variants by a single method for $c^N$ with all subtasks in mode $N$.
In its $F$-variants, remove the check and its orderings but retain the marking.
This preserves the equality of \cref{thm:main}, total order, acyclicity, tail recursion and regularity, without increasing size.
\end{proposition}

\begin{proofidea}
The new $N$-method has no marked subtask and hence no subtask in mode $F$ or $R$, so every task in mode $F$ and every release still lies on a marked path below an opening check, and \cref{lem:window} applies unchanged.
For completeness, open each window at the first application with a non-empty precondition among those sharing a first action.
\end{proofidea}

\paragraph{Lifted domains.}
Since marks select subtask occurrences and $\free$ and $\lock$ are nullary, \cref{def:comp} applies unchanged to lifted methods and commutes with grounding.
Each check takes the method parameters occurring in its precondition, and with \cref{prop:trivial} the result is ordinary HDDL without method preconditions.

\paragraph{Complexity.}
By \cref{thm:main,prop:size}, plan existence under the first-action semantics reduces polynomially to plan existence without method preconditions within the acyclic, tail-recursive, regular and totally ordered fragments; since problems without method preconditions are a special case, the tight bounds of \citet{ErolHN96} and \citet{AlfordBA15} carry over.
The totally ordered and regular cases already follow from \citet{ErolHN96}, whose state constraints can express method preconditions, but the acyclic and tail-recursive cases are new.

\section{Discussion and Related Work}\label{sec:discussion}

\paragraph{Related work.}
Besides the SHOP family and UMCP \citep{ErolHN94}, the first-action reading underlies the before-constraints of the textbook formalism \citep[Ch.~11]{GhallabNT04} and of parsing-based verifiers \citep{BartakMC18,BartakOMBB20,OndrckovaBBB22,LinBOBB23}, whereas compilation-based verifiers follow HDDL \citep{BehnkeHB17,HoellerWBB22}.
Unlike \citet{OndrckovaPB24}, we keep empty methods as no-ops and remove only their preconditions, and like \citet{HoellerWBB22}, we use state to restrict interleavings.
Conversely, $\Lhddl(P) = h(\Lfirst(H(P)))$ because $H(P)$ has no method preconditions, so the two readings are polynomially inter-compilable.
\citet{HoellerBBBPA20} argue that method preconditions usually serve as search advice that should be excluded from a model.
Checking such advice too early can admit a method whose precondition no longer holds when its first action executes.
Method preconditions also bind variables for later subtasks and constrain the start of a method.

\paragraph{Extensions and open questions.}
SHOP2's methods with several precondition--network pairs compile exactly by creating one method per pair with the earlier preconditions negated, since all pairs are tested in the same state.
One open question is whether a similar compilation exists when an ancestor's precondition must hold before its first action that is not a no-op (\cref{sec:prelims}).
Another is how to extend the lock to before-constraints on subsets of subtasks \citep{OndrckovaB23} and to after- and between-constraints, where constraints on the same method interact.
We also leave open whether quadratic size is necessary.

\section{Conclusion}\label{sec:conclusion}

We have shown that the first-action semantics of method preconditions admits a compilation into standard partially ordered HTN planning.
This resolves a question left open since HDDL was introduced, and makes existing HDDL planners and complexity results applicable to the semantics of the SHOP family.
Allowing HDDL domains to declare their intended semantics would make this distinction explicit.

\section*{Declaration of AI Use}
The author used Anthropic's Claude interactively to assist in developing and refining proof arguments, preparing an initial manuscript draft, and formalising the results in Lean.
The central compilation approach was conceived independently by the author, who reviewed and verified the AI-assisted work and takes full responsibility for the paper's contents.

\bibliography{references}

\appendix
\crefalias{section}{appendix}
\section{Proofs for \Cref{sec:semantics}}\label{app:semantics}

Several proofs group method applications by their first action.
Fix a complete decomposition and a linearisation $a_1,\ldots,a_\ell$ of its final network, and let $A_i$ be the set of applications $\mu$ with $\first(\mu)=i$.

\begin{restate}{Observation}
Every $A_i$ is either empty or a contiguous segment $\mu_0,\ldots,\mu_k$ of the ancestor path of $a_i$, ordered from ancestor to descendant, whose last element $\mu_k$ is the parent of $a_i$.
\end{restate}
\begin{proof}
Every $\mu \in A_i$ has $a_i$ in its yield, so it lies on the ancestor path of $a_i$.
If $\mu \in A_i$ and $\nu$ lies between $\mu$ and $a_i$ on this path, then the yield of $\nu$ contains $a_i$ and is contained in the yield of $\mu$, so $\first(\nu)=i$.
In particular, the parent of $a_i$ is in $A_i$ whenever $A_i$ is non-empty.
\end{proof}

\begin{restate}{\Cref{lem:insert} (Check insertion)}
Take a complete decomposition of $P$, a linearisation $a_1,\ldots,a_\ell$ of its final network, and the corresponding decomposition of $H(P)$.
Placing each check $t_\mu$ directly before $a_{\first(\mu)}$, with the checks placed before the same action ordered from ancestor to descendant, gives a linearisation of the final network of $H(P)$.
\end{restate}
\begin{proof}
By the observation, the order within each block of checks is well defined.
Constraints between actions are those of the given linearisation, so we only consider constraints on a check $t_\mu$, where $\mu$ is applied to a task $u$.
In $H(P)$, $t_\mu$ precedes the subtasks of $\mu$, and inherits every constraint of $u$ and its ancestors.

\emph{Descendants of $\mu$.}
Every action in the yield of $\mu$ is placed at or after $a_{\first(\mu)}$, and hence after $t_\mu$.
A descendant application $\mu'$ has a smaller yield, so $\first(\mu') \geq \first(\mu)$, and its check is placed in a later block or after $t_\mu$ in the same block.

\emph{Inherited constraints.}
Let $u'$ be $u$ or one of its ancestors, and let $u' \prec v$.
Every action below $v$, and every check below $v$, is placed at or directly before an action below $v$, while $t_\mu$ is placed directly before an action below $u'$.
In the given linearisation, every action below $u'$ precedes every action below $v$, so $t_\mu$ precedes every occurrence below $v$.
The case $v \prec u'$ is symmetric.
\end{proof}

\begin{restate}{\Cref{prop:compare}}
$\Lfirst(P) \subseteq \Lhddl(P)$.
The two languages coincide if $P$ is totally ordered, and the inclusion can be strict otherwise.
\end{restate}
\begin{proof}
\emph{Inclusion.}
Place the checks of a first-action solution as in \cref{lem:insert}.
Checks do not change the state, so each $t_\mu$ is executed in $s_{\first(\mu)-1}$, which satisfies $\pre(\mu)$.
This gives a solution of $H(P)$ with the same action word.

\emph{Total order.}
Take a solution of $H(P)$ and an application $\mu$ to a task $u$, and let $x$ be an occurrence between $t_\mu$ and the first action of $\mu$.
Since $P$ is totally ordered, every occurrence that is not below $u$ is ordered with respect to $u$, and so precedes $t_\mu$ or follows the whole yield of $\mu$.
Hence $x$ is below $u$, and since it precedes the first action of $\mu$, it is a check.
Checks do not change the state, so $\pre(\mu)$ holds directly before the first action of $\mu$, and erasing the checks gives a first-action solution with the same word.

\emph{Strictness.}
Linearisation (i) of the running example is in $\Lhddl(P)$ but not in $\Lfirst(P)$.
\end{proof}

\begin{restate}{\Cref{prop:shop} (SHOP2)}
For a ground problem $P$, the action words of the runs of SHOP2's task-selection rule are exactly $\Lfirst(P)$.
\end{restate}
\begin{proof}
With the no-op normalisation, removing a task by an empty method is a decomposition followed by the execution of its no-op, so a run consists of decompositions and executions.

\emph{From runs to first-action solutions.}
A run only chooses minimal tasks, so its actions form a linearisation of the final network.
After decomposing a task by an application $\mu$, the run stays below $\mu$ until it executes an action $a$.
No action of $\mu$'s yield was executed earlier, so $a$ is the first action of $\mu$.
The state does not change in between, so $\pre(\mu)$ holds directly before $a$.

\emph{From first-action solutions to runs.}
For $i = 1,\ldots,\ell$, decompose the tasks of $A_i$ from the top down and then execute $a_i$.
By the observation, each of these decompositions after the first chooses a subtask of the previous one, as SHOP2 requires, and $a_i$ is a subtask of the last; if $A_i$ is empty, $a_i$ follows an execution, after which SHOP2 may choose any minimal task.
The topmost task of $A_i$ is available, because its parent application has an earlier first action and was therefore applied earlier.
Every chosen task is minimal, since a remaining predecessor would contain an action that must precede $a_i$ but has not been executed.
Each precondition is evaluated in $s_{i-1}$, which satisfies it by \cref{def:imm}.
\end{proof}

\section{Proofs for \Cref{sec:correctness}}\label{app:correctness}

\begin{restate}{\Cref{lem:chains} (Marked chains)}
In a complete decomposition of $\Comp_{\pc}(P)$, each application to a task in mode $N$ begins a unique marked chain.
The chain passes through zero or more applications to tasks in mode $F$ and ends at one action occurrence in mode $R$.
If the chain's method applications are $\mu_0,\ldots,\mu_k$ and its release is $r$, then
\begin{equation}
  \pc_{\mu_0}^N \prec \pc_{\mu_1}^F \prec \cdots
  \prec \pc_{\mu_k}^F \prec r
\end{equation}
in the final network.
Every check occurrence and every release occurrence belongs to exactly one marked chain.
\end{restate}
\begin{proof}
Every compiled method has exactly one marked subtask.
Starting from an application to a task in mode $N$, follow marked subtasks: each compound one is in mode $F$, and the method applied to it again has one marked subtask.
Since the decomposition is finite, the path ends at a primitive marked subtask, which is in mode $R$.

Conversely, a task in mode $F$ or $R$ is the marked subtask of its parent application, whose task is in mode $N$ or $F$.
Since the initial tasks are in mode $N$, following parents upwards reaches an application in mode $N$.
Hence every application to a task in mode $F$ and every release lies on exactly one chain, and so does every check.

Each check precedes all subtasks of its application and hence all of their descendants, including the next check of the chain and its release, which gives \eqref{eq:chain-order}.
\end{proof}

\begin{restate}{\Cref{lem:window} (Lock ownership)}
In every executable linearisation of a complete decomposition of $\Comp_{\pc}(P)$, each lock window is exactly
\(
  \pc_{\mu_0}^N,\;\pc_{\mu_1}^F,\;\ldots,\;\pc_{\mu_k}^F,\;r
\)
for one marked chain of \cref{lem:chains}.
In particular, no window contains an action, check or release from outside its chain.
\end{restate}
\begin{proof}
A chain is \emph{unopened} before its opening check is executed, \emph{active} until its release is executed, and \emph{completed} afterwards.
We show by induction along the linearisation that at most one chain is active, and that $\lock$ holds exactly when some chain is active and $\free$ exactly when none is.
This holds initially, and only opening checks and releases change $\free$ and $\lock$.

If no chain is active, $\lock$ is false, so the next occurrence is a normal action, which changes nothing, or an opening check, which activates its chain and swaps $\free$ for $\lock$.
If a chain $C$ is active, $\free$ is false, so the next occurrence is an inner check or release of some chain $C'$.
By \eqref{eq:chain-order}, $C'$ cannot be unopened, since its opening check precedes it, and cannot be completed, since its release follows it and occurs once.
So $C' = C$; an inner check changes nothing, and the release completes $C$ and restores $\free$.

Hence the window opened by $\pc_{\mu_0}^N$ contains only occurrences of its own chain and ends at its release.
By \eqref{eq:chain-order} and since every occurrence is executed, it contains all inner checks of the chain in chain order.
\end{proof}

\begin{restate}{\Cref{thm:main} (Soundness and completeness)}
For every ground problem $P$,
\(
  h\bigl(\Lang(\Comp_{\pc}(P))\bigr)=\Lfirst(P).
\)
\end{restate}
\begin{proof}
\emph{Soundness.}
Take a solution of $\Comp_{\pc}(P)$, and let $a_1,\ldots,a_\ell$ be the linearisation of the underlying decomposition of $P$ obtained by erasing checks and annotations.
It respects the ordering constraints of $P$, and it is executable because each copy $o^N$ or $o^R$ acts on $F$ like $o$ and checks do not change $F$.

Let $\mu$ be a method application and $r$ the release of its chain, which lies in the yield of $\mu$.
The check of $\mu$ precedes every action in the yield of $\mu$, and by \cref{lem:window} only checks occur between this check and $r$.
So $r$ is the first action of $\mu$, and $\pre(\mu)$ holds directly before $r$ because it held at the check.
Hence we have a first-action solution with the projected word.

\emph{Completeness.}
Take a first-action solution with linearisation $a_1,\ldots,a_\ell$.
In each application $\mu$, mark the subtask below which $a_{\first(\mu)}$ lies.
This subtask is minimal, since a predecessor would contain an action that must precede $a_{\first(\mu)}$.
The marks determine a decomposition of $\Comp_{\pc}(P)$, whose networks have the same shape as those of $H(P)$, so placing the checks as in \cref{lem:insert} gives a linearisation of its final network.

By the observation, each $a_i$ with $A_i$ empty is in mode $N$ with no checks before it, since its parent does not mark it.
Otherwise, $A_i = \mu_0,\ldots,\mu_k$ ends at the parent of $a_i$, which marks $a_i$, and each $\mu_j$ with $j \geq 1$ is marked by $\mu_{j-1}$.
The task of $\mu_0$ is in mode $N$, since its parent, if any, is not in $A_i$ and so does not mark it.
So the block before $a_i$ is $\pc^N,\pc^F,\ldots,\pc^F,a_i^R$.
The linearisation therefore consists of single normal actions and complete windows, so the lock facts allow every step and $\free$ holds between them.
The checks are executed in $s_{\first(\mu)-1}$, which satisfies $\pre(\mu)$, and the actions affect $F$ as in the given solution, so the linearisation is executable and projects under $h$ to the given word.
\end{proof}

\begin{restate}{\Cref{cor:bijection}}
Identify solutions by their decomposition forest and linearisation.
Then erasing checks and annotations is a bijection from the solutions of $\Comp_{\pc}(P)$ to the first-action solutions of $P$.
\end{restate}
\begin{proof}
The map is well defined and onto by \cref{thm:main}.
For injectivity, the soundness proof shows that in a compiled solution the release below each application $\mu$ is its first action, so the marks, and hence the modes, are determined by the image.
By \cref{lem:window}, each check lies in its chain's window in chain order, directly before the release, so the positions of the checks are determined as well.
\end{proof}

\section{Proofs for \Cref{sec:size}}\label{app:size}

\begin{restate}{\Cref{prop:size}}
Let $k_m=|\mathrm{Min}(T_m,\prec_m)|$ be the number of minimal subtasks of method $m$.
The compilation generates $2|N_C|$ compound task names, $2|N_P|+2|M|$ primitive task names and $2\sum_{m\in M}k_m$ methods.
With explicit task and ordering lists, its description size is
\(
  O(|P|+\sum_{m\in M}k_m(|T_m|+|\prec_m|)).
\)
It can be constructed in polynomial time, has size $O(|P|^2)$ in general and $O(|P|)$ if every method is totally ordered, and preserves total order, acyclicity, tail recursion and regularity.
\end{restate}
\begin{proof}
Each method has two variants per minimal subtask, each with $|T_m|+1$ tasks and $|\prec_m|+|T_m|$ ordering pairs, and all other components have size $O(|P|)$.
This gives the bound, and a construction time proportional to it.
Since $k_m\leq |T_m|$, the bound is quadratic, which a method with $n$ unordered primitive subtasks attains; under total order $k_m=1$, giving a linear bound.
The check is primitive and precedes all subtasks, so total order is preserved and a last subtask stays last.
Giving $c^N$ and $c^F$ the level of $c$, each compiled method satisfies the tail-recursion and regularity conditions of its original, and each compiled dependency between compound tasks projects to an original one, which preserves acyclicity.
\end{proof}

\begin{restate}{\Cref{prop:trivial} (Methods without preconditions)}
For every method $m$ with $\pre(m)=\emptyset$, replace its $N$-variants by a single method for $c^N$ with all subtasks in mode $N$.
In its $F$-variants, remove the check and its orderings but retain the marking.
This preserves the equality of \cref{thm:main}, total order, acyclicity, tail recursion and regularity, without increasing size.
\end{restate}
\begin{proof}
\emph{Soundness.}
The new $N$-method has no marked subtask, so it starts no chain and creates no task in mode $F$ or $R$.
Every task in mode $F$ and every release therefore still lies on a marked path below an opening check, with the checks of the path ordered as in \eqref{eq:chain-order}.
The proof of \cref{lem:window} then applies unchanged, so every non-empty precondition holds directly before its method's first action.

\emph{Completeness.}
Let $\mu_0,\ldots,\mu_k$ be the applications with first action $a$ in a first-action solution.
If all have empty preconditions, use their $N$-variants.
Otherwise, let $\mu_j$ be the first with a non-empty precondition; use $N$-variants above $\mu_j$, open the window at its check, and pass the mark down through $F$-variants to $a^R$.
As in \cref{thm:main}, the remaining checks and the release form an executable window.
Removing checks and variants cannot increase size or affect the structural properties.
\end{proof}

\end{document}